\documentclass[10pt]{article}
\usepackage{color, palatino}

\usepackage{bbm}
\usepackage{mathptmx}       
\usepackage{helvet}         
\usepackage{courier}        
\usepackage{type1cm}        
\usepackage{makeidx}         
\usepackage{graphicx}        
\usepackage{multicol}        
\usepackage[bottom]{footmisc}
\usepackage{amsfonts,amssymb,mathrsfs,amsmath,graphicx}

\newtheorem{theorem}{Theorem}

\newtheorem{definition}{Definition}

\newtheorem{remark}{Remark}
\newtheorem{example}{Example}

\newcommand{\new}{\newcommand}

\new{\R}{\mathbb R}
\new{\C}{\mathbb C}
\new{\N}{\mathbb N}
\new{\regr}{g}
\new{\hh}{\mathcal H}
\new{\Reg}{\mathcal{R}}
\new{\kk}{\mathcal K}
\new{\Ly}{\mathcal{L}(\Y)}
\new{\Fy}{\mathcal{F}(\X,\Y)}
\new{\CbX}{\mathcal{C}(\X;\Y)}
\new{\la}{\lambda}
\new{\eps}{\epsilon}
\new{\empp}{{\mathcal{E}_{{\bf z}_n}}}
\new{\empo}{{\hat{\mathcal{E}}_{{\bf Z}_n(\omega)}}}
\new{\emp}{{\hat{\mathcal{E}}_{\ts}}}
\new{\err}{{\mathcal{E}}_\rho}
\new{\er}{{\mathcal{E}}}
\new{\DD}{\mathcal{D}}
\new{\ud}{\mathrm{d}}
\new{\beeq}[2]{\begin{equation}\label{#1}#2\end{equation}}
\new{\ran}[1]{\operatorname{Ran}#1}
\new{\Ker}[1]{\operatorname{Ker}#1}
\new{\tr}[1]{\operatorname{Tr}#1}
\new{\supp}[1]{\operatorname{supp}#1}

\new{\lspan}[1]{\operatorname{span}\{#1\}}
\new{\lspanc}[1]{\overline{\operatorname{span}}\{#1\}}
\new{\argmin}{\operatornamewithlimits{argmin}}

\new{\PP}[1]{{\mathbb P}\left(#1\right)}
\new{\PPd}{\mathbb P}
\new{\EE}[1]{{\mathbb E}\left[#1\right]}

\new{\subg}{\tilde{\nabla}}
\new{\prox}{\mathrm{prox}}
\new{\fat}{\mathrm{fat}}
\new{\inff} {\inf_{f\in {\F}_{Z,L}}\err(f)}
\new{\FZL}{{\F}_{Z,L}}
\new{\X}{\mathcal{X}}
\new{\Y}{\mathcal{Y}}
\new{\Z}{\mathcal{Z}}
\new{\F}{\mathcal {F}}
\new{\A}{\mathcal{A}}
\new{\um}{1,\ldots,m}
\new{\tv}{T}
\new{\trs}{{{\bf z}_n}}
\new{\tsi}{{{\bf Z}^i_n}}
\new{\tsmi}{{\bf Z}^{\setminus i}_n}
\new{\vz}{n}
\new{\rhox}{\rho_{\X}}
\new{\ro}{\rho_{_{\!Z}}}
\new{\Ldue}{L^2(X,\rhox)}
\new{\f}{f_\rho}
\new{\fh}{f_\hh}
\new{\dd}{\mathrm{d}}
\new{\IK}{\Phi_{\rho}}
\new{\Sx}{\Phi_n}
\new{\vy}{\mathbf y}
\new{\K}{\mathbf K}
\new{\ik}{i_\nu}
\new{\LK}{L_\nu}

\begin{document}

\title{On Learnability, Complexity and Stability}

\author{\normalsize{ Silvia Villa$^{\dagger}$, Lorenzo Rosasco$^{\dagger,\S}$, Tomaso Poggio$^{\top,\dagger}$}\\
\small \em $\top$  
CBCL,  Massachusetts Institute of Technology\\
\small \em $\dagger$ 
LCSL , Istituto Italiano di Tecnologia  and Massachusetts Institute of Technology\\
\small \em $\S$ 
DIBRIS, Universita' degli Studi di Genova\\
\small \tt  silvia.villa@iit.it, \{lrosasco,tp\}@mit.edu}

\maketitle


\maketitle

\begin{abstract}
We consider the fundamental question of learnability of a hypotheses class in the supervised learning setting and in the general
learning setting introduced by Vladimir Vapnik.  We survey classic results characterizing  learnability in term of
suitable notions of complexity, as well as more recent results that establish  the  connection between learnability and
stability of a learning algorithm.
\end{abstract}

\section{Introduction}

A key question  in  statistical learning is  which hypotheses (function) spaces are learnable.
Roughly speaking, a hypotheses space is learnable  if   there is a  consistent learning algorithm, i.e. one 
returning an optimal solution    as the number of sample goes to infinity. Classic results  for   supervised  learning 
  characterize learnability of a function class in terms of its complexity (combinatorial dimension)  \cite{VapChe71,Vap95,AloBenCes97,AntBar99,KeaSch94,BarLonWil96}.  
  Indeed,  minimization of the empirical risk  on a  function class having finite complexity can be shown to be consistent.  A key aspect in this approach is the connection with empirical process theory results showing that finite combinatorial dimensions characterize  function   classes for which a uniform law of large numbers holds, namely uniform Glivenko-Cantelli classes \cite{DudGinZin91}.

More recently,  the concept of stability has emerged  as an alternative and effective method to design consistent learning algorithms \cite{bousquet02stability}. 
Stability  refers  broadly to continuity properties of  learning algorithm to its input and it is known to play a crucial role in 
in  regularization theory \cite{enhane}. 
Surprisingly, for certain classes of loss functions, a suitable notion of stability of  ERM  
can be shown to characterize learnability of a function class \cite{KutNiy02,PogRifMuk04,MukNyiPog06}. 

In this paper, after recalling  some basic concepts (Section 2), we review 
results characterizing learnability in terms of complexity and stability in  supervised learning (Section 3)
and in the so called general learning  (Section 4). We conclude with some remarks and open questions.

\section{Supervised Learning, Consistency and Learnability}\label{sec:uno}

In this section, we introduce  basic concepts  in Statistical Learning Theory (SLT). First, we describe the supervised learning setting, and then, define the notions of  consistency of a learning algorithm and of learnability of a hypotheses class. 

Consider a probability space $(\Z, \rho)$, where  $\Z=\X\times\Y$, with  $\X$ a measurable  space and $\Y$ a closed subset of $\R$.  A loss function  is  a measurable map   $\ell: \R\times\Y\to [0,+\infty)$. We are interested in the 
problem of  minimizing the expected risk,
\beeq{eq:exp_risk}
{\inf_{\F}\err,\quad\quad \err(f)=\int_{\X\times\Y} \ell(f(x),y)\,d\rho(x,y),}
where $\F\subset \Y^\X$ is the set of measurable functions 
from $\X$ to $\Y$ (endowed with the product topology and the corresponding Borel $\sigma$-algebra). 
The probability distribution $\rho$ is assumed to be  fixed but known only through a training set, i.e. a set of   pairs ${\bf z}_n=((x_1,y_1),\ldots,(x_n,y_n))\in \Z^n$ sampled  identically and independently  according  to $\rho$. 
Roughly speaking, the problem of supervised learning  is that of  approximatively 
solving Problem~\eqref{eq:exp_risk} given a training set ${\bf z}_n$.  
\begin{example}[Regression and Classification]
In (bounded) regression  $\Y$ is a bounded interval in  $\R$, while in  binary classification $\Y=\{0,1\}$. 
Examples of  loss functions are the  square loss $\ell(t,y)=(t-y)^2$ in regression and the 
misclassification loss $\ell(t,y)=\mathbbm{1}_{\{t\neq y\}}$ in classification. 
See \cite{Vap95} for a more exhaustive list of loss functions. 
\end{example}
In the next section, the notion  of approximation considered in SLT is defined rigorously.  
We first introduce the concepts of hypotheses space and learning  algorithm. 
\begin{definition}  A {\em hypotheses space} is a set of functions  $\hh\subseteq \F$. 
We say that $\hh$ is {\em universal} if 
 $
\inf_\F\err =\inf_\hh\err,
$
for all distributions $\rho$ on $\Z$.
\end{definition}
\begin{definition}\label{Lalg}  A {\em learning algorithm} $A$ on $\hh$ is a  map,
\[
A: \ \bigcup_{n\in\mathbb{N}} \Z^n \to \hh,\quad \ {\bf z}_n\mapsto A_{{\bf z}_n}=A({\bf z}_n),
\] 
such that, for all $n\ge 1$,  $A_{|\Z^n}$ is measurable with respect to the completion of the product $\sigma$-algebra on $\Z^n$.
\end{definition}
Empirical Risk Minimization (ERM) is arguably the most popular  example 
of learning algorithm in SLT.
\begin{example} Given  a training set ${\bf z}_n$  the empirical risk  $\empp:\F \to \mathbb{R}$ is defined as 
\[
\empp(f) =\frac{1}{n}\sum_{i=1}^n\ell(f(x_i),y_i).
\]
Given a hypotheses space $\hh$,  ERM  on $\hh$  is defined by   minimization of the empirical risk on $\hh$.
\end{example}
We add one  remark.
\begin{remark}[ERM and Asymptotic ERM]\label{rem:aerm}
In general some care is needed while defining ERM since a (measurable) minimizer might not be ensured
to exist.  When $\Y=\{0,1\}$ and $\ell$ is the misclassification loss function, it is easy to see that a minimizer 
exists (possibly non unique). In this case measurability is studied for example in Lemma 6.17 in \cite{SteChr08}. When considering 
more general loss functions or regression problems one might need to consider learning algorithms 
defined by suitable (measurable) almost minimizers of the empirical risk (see e.g. Definition~\ref{AERM}).
\end{remark}


%
\subsection{Consistency and  Learnability}


Aside from computational considerations, the  following definition formalizes in which sense a 
learning algorithm  approximatively solves Problem~\eqref{eq:exp_risk}.
\begin{definition}\label{def:ucons} We say that a learning algorithm $A$ on $\hh$ is {\em  uniformly consistent}\footnote{Consistency can de defined with respect to other   convergence notions   for random variables. If the loss function is bounded, convergence in probability is equivalent to convergence in expectation.} if  
\[
\forall \epsilon>0,\quad
\lim_{n\to+\infty} \sup_{\rho} \rho^n{ \big(\{\trs\,:\err(A_\trs)- \inf_\hh\err>\epsilon\}\big) } = 0,\]
and  {\em universally uniformly consistent} if $\hh$ is universal. 
\end{definition}
The  next definition  shifts the focus  from  a learning algorithm 
on $\hh$, to  $\hh$ itself. 
\begin{definition}\label{def:ucons} We say that a  space $\hh$ is {\em uniformly learnable} if there exists a uniformly consistent learning algorithm on $\hh$.  If  $\hh$ is also universal we say that it is   {\em universally  uniformly learnable}.
\end{definition}
Note that, in the above definitions,  the term ``uniform" refers to the distribution for which consistency holds, whereas ``universal" refers to the possibility of solving Problem~\eqref{eq:exp_risk} without a bias due to the choice of $\hh$.
The requirement of uniform learnability implies the existence of a learning rate for $A$ \cite{SteChr08} or equivalently
 a bound on the sample complexity \cite{AntBar99}.  
The following classical result, sometimes called the "no free lunch" theorem, shows that uniform universal learnability of a  hypotheses space  is too much to hope for.

\begin{theorem}\label{thm:nfl}  Let $\Y=\{0,1\}$, 
and $\X$  such that there exists a measure  $\mu$ on $\X$ having an atom-free distribution. 
Let $\ell$ be the misclassification loss.  If $\hh$ is universal, then $\hh$ is not uniformly learnable.  
\end{theorem}
The proof of the above result is based on Theorem 7.1 in \cite{DevGyoLug96}, 
which shows that for each learning algorithm $A$ on $\hh$ and any fixed $n$, there exists a measure $\rho$ on  $\X\times\Y$  
such that the expected value of $\err(A_{{\bf z}_n})-\inf_\hh\err$ is greater than $1/4$. 
A  general form of the no free lunch theorem, beyond   classification, is given  in \cite{SteChr08} (see Corollary 6.8).
In particular, this  result shows that  the no free lunch theorem holds for  convex loss functions, as soon as  there are two probability distributions $\rho_1,\rho_2$ such that $\inf_\hh \er_{\rho_1}\neq \inf_\hh \er_{\rho_2}$  (assuming that minimizers exist). Roughly speaking, if there exist two learning problems with distinct  solutions, then $\hh$ cannot be universal uniformly learnable (this latter condition becomes more involved when the loss is not convex).

The no free lunch theorem shows that universal uniform consistency is  too strong of a requirement. Restrictions on either the class of considered distributions $\rho$ or the hypotheses spaces/algorithms are needed to define a meaningful problem.  In the following,  we will follow the latter approach  where assumptions on $\hh$ (or $A$), but not on the class distributions $\rho$, are made. 

\section{Learnability of a Hypotheses space}

In this section we study uniform learnability by putting appropriate restrictions on the hypotheses space $\hh$. We are interested in conditions which are not only sufficient
but also necessary. 
We discuss two series of results. The first is classical and characterizes learnability of a hypotheses space in terms of suitable complexity measures. 
The second, more recent, is based on the stability (in a suitable sense) of  ERM on $\hh$. 

\subsection{Complexity and Learnability}

 Classically assumptions on $\hh$ are imposed in the form of restrictions on its  
"size" defined in terms of suitable notions of combinatorial dimensions (complexity).
The following definition of complexity for a class of binary  valued functions has been introduced in \cite{VapChe71}.
\begin{definition}\label{def:VCdim} Assume $\Y=\{0,1\}$.  We say that $\hh$ {\em shatters} $S\subseteq \X$ if for each $E\subseteq S$ there exists $f_E\in \hh$ such that 
$ f_E(x)=0$, if $x\in E$, and $f_E(x)=1$ is $x\in S\setminus E$.
The {\em VC-dimension} of $\hh$ is defined as 
\[
\mathrm{VC}(\hh)=\max\{d\in \mathbb{N}\,:\,\exists\, S=\{x_1,\ldots x_d\} \text{ shattered by } \hh\}
\]
\end{definition}
The VC-dimension  turns out to be related 
to a special class of functions, called uniform Glivenko-Cantelli, for which 
a uniform form of the law of large numbers holds \cite{DudGinZin91}.  
\begin{definition}\label{def:uGC}
We say that $\hh$ is a {\em uniform Glivenko-Cantelli (uGC) class} if it has the following property 
\[
\forall \epsilon>0, \quad \lim_{n\to +\infty} \sup_{\rho} \rho^n\Big(\Big\{\trs\,:\,\sup_{f\in\hh}\left \vert\err(f)-\empp(f)\right \vert>\epsilon \Big\}\Big)=0\,.
\]
\end{definition}
The following  theorem completely characterizes learnability in classification.
\begin{theorem}\label{thm:vapche} Let $\Y=\{0,1\}$  
and $\ell$ be the misclassification loss. Then the following conditions are equivalent:
\begin{enumerate}
\item  $\hh$ is uniformly learnable,
\item  ERM  on $\hh$ is uniformly consistent,
\item $\hh$ is a uGC-class,
\item the $VC$-dimension of  $\hh$ is finite.
\end{enumerate} 
\end{theorem}
The proof of the above result can be found for example in  \cite{AntBar99} (see Theorems 4.9, 4.10 and 5.2). 
The characterization of uGC classes  in terms of  combinatorial dimensions is a central theme in empirical process theory \cite{DudGinZin91}.  The results  on binary valued functions are  essentially due to Vapnik and Chervonenkis \cite{VapChe71}.  The proof that uGC of $\hh$ implies its learnability 
is  straightforward. The key step in the above proof is showing that 
learnability is sufficient for  finite VC-dimension, i.e.  $\mathrm{VC}(\hh)<\infty$.
The proof of this last step crucially depends on the considered loss function. \\
A similar result holds for bounded regression with the square \cite{AloBenCes97,AntBar99} and absolute loss functions \cite{KeaSch94,BarLonWil96}.   In this case, a new notion of complexity needs to be defined since 
the $VC$-dimension of real valued function classes is  not defined.
Here, we recall the definition of $\gamma$-fat shattering dimension of a class of functions $\hh$ originally introduced in \cite{KeaSch94}.

\begin{definition} Let $\hh$ be a set of functions from $\X$ to $\mathbb{R}$ and $\gamma>0$. Consider  $S=\{x_1,\ldots,x_d\}\subset \X$. Then $S$ is $\gamma$-{\em shattered by} $\hh$ if there are real numbers $r_1,\ldots,r_d$ such that for each $E\subseteq S$ there is a function $f_E\in\hh$ satisfying
\[
\begin{cases}f_E(x) \leq r_i-\gamma & \forall x\in S\setminus E\\
f_E(x)\geq r_i+\gamma & \forall x\in E.
\end{cases}
\] 
We say that $(r_1,\ldots, r_d)$ witnesses the shattering. The $\gamma$-{\em fat shattering dimension of} $\hh$ is 
\[
\fat_\hh(\gamma)=\max\{d\,:\, \exists\, S=\{x_1,\ldots,x_d\}\subseteq \X \text{ s.t. $S$ is $\gamma$-shattered by $\hh$}\}. 
\]
\end{definition}

As mentioned above, an analogous of Theorem  \ref{thm:vapche} can be proved for bounded regression with the square and absolute losses, if condition 4) is replaced by $\fat_\hh(\gamma)<+\infty$ for all $\gamma>0$.  
We end noting that  is an open question proving that the above results holds for loss function other than the square and absolute loss.
%

\subsection{Stability and Learnability}

In this section we show that learnability of a hypotheses space $\hh$ is equivalent to the stability (in a suitable sense) of ERM on $\hh$. It is useful to introduce the following notation. For a given loss function $\ell$, let $L:\F\times Z \to [0,\infty)$ be defined as 
$L(f, z)=\ell(f(x),y)$, for $f\in \F$ and $z=(x,y)\in \Z$. Moreover, let  ${\bf z}_n^{ i}$ be the training $\trs$ with the $i$-th point removed.   With the above notation, the relevant notion of stability is given by the following definition.

\begin{definition}\label{def:cCVloo} A learning algorithm $A$ on $\hh$  is  {\em uniformly  $CV_{loo}$ stable} if there exist  sequences $(\beta_n,\delta_n)_{n\in\mathbb{N}}$ such that $\beta_n\to 0$, $\delta_n\to 0$ and 
\begin{equation}\label{CVloo}
\sup_{\rho} \rho^n{\{ |L(A_{{\bf z}_n^{ i}},z_i)-L(A_{{\bf z}_n},z_i)| \leq \beta_n\}}\geq 1-\delta_n\,,
\end{equation}
for all  $i\in\{1,\ldots,n\}$.
\end{definition}
Before illustrating the implications of the above definition to learnability we first add a few comments and historical remarks. 
We note that, in a broad sense, stability refers to a quantification of the continuity of a map with respect to its input.  The key  role of  stability in  learning has long been advocated on the basis of the interpretation of supervised learning as an ill-posed inverse problems \cite{MukNyiPog06}. 
Indeed, the concept of stability is central in the  theory of regularization of  ill-posed problem \cite{enhane}. A first quantitative 
connection between the performance 
of a symmetric learning algorithm\footnote{We say that a learning algorithm $A$ is symmetric if it  does not depend on the order of the points in ${\bf z}_n$.} and a notion of stability is  derived in the seminal paper \cite{bousquet02stability}. 
Here a 
notion of stability, called uniform stability, is shown to be sufficient for consistency.
If we let  ${\bf z}_n^{ i,u}$ be the training $\trs$ with the $i$-th point replaced by $u$,
uniform stability is defined as,
\begin{equation}\label{ustab}
|L(A_{{\bf z}_n^{ i,u}},z)-L(A_{{\bf z}_n},z)| \leq \beta_n,
\end{equation}
 for all $\trs\in \Z^n$, $u,z\in \Z^n$  and $i\in\{1,\ldots,n\}$.
 A thorough investigation of  weaker notions of stability  is given in \cite{KutNiy02}. 
Here, many different notions of stability are shown to be sufficient for consistency (and learnability) and the question is raised of whether stability (of ERM on $\hh$) can be shown to be necessary for learnability of $\hh$.  In particular a definition of $CV$ stability for ERM  is shown to be  necessary and sufficient for learnability
 in a Probably Approximate Correct  (PAC) setting,  that is when 
$\Y=\{0,1\}$ and for some $h^*\in \hh$, $y=h^*(x)$,  for all $x\in \X$.
Finally,  Definition~\ref{def:cCVloo} of $CV_{loo}$ stability is given  and studied in \cite{MukNyiPog06}.
When compared to uniform stability, we see that: 1) the   ``replaced one'' training set ${\bf z}_n^{ i,u}$ is considered instead of the  ``leave one out'' training set ${\bf z}_n^{ i}$; 2) the error is evaluated on the point $z_i$
which is left out, rather than any possible $z\in \Z$; finally 3) the condition is assumed to hold 
for  a fraction $1-\delta_n$ of training sets (which becomes increasingly larger as $n$ increases) rather than uniformly for any training set $\trs\in \Z^n$.

The importance of $CV_{loo}$ stability is made clear  by the following result.

%
%
%
%

\begin{theorem} \label{stabtheo}
Let $\Y=\{0,1\}$ and $\ell$ be the misclassification loss function. 
Then the following conditions are equivalent,
\begin{enumerate}
\item $\hh$ is uniformly learnable,
\item  ERM  on $\hh$ is $CV_{loo}$ stable  
\end{enumerate}
\end{theorem}
The proof of  the above result is given in  \cite{MukNyiPog06} and 
 is based on essentially two steps. The first is  proving
that  $CV_{loo}$ stability of ERM on $\hh$ implies that ERM is uniformly consistent. 
The second is showing that if $\hh$ is a uGC class then  ERM on $\hh$ is $CV_{loo}$ stable. 
Theorem~\ref{stabtheo} then follows from Theorem~\ref{thm:vapche} (since uniform consistency of ERM on $\hh$ and $\hh$ being uGC are equivalent). 

Both steps in the above proof can be generalized to regression 
as long as the loss function is assumed to be  bounded. 
The latter assumption holds for example if the loss function satisfies a suitable  Lipschitz condition
 and 
$\Y$  is compact  (so that   $\hh$ is  a set of uniformly bounded functions).     
However, generalizing Theorem~\ref{stabtheo} beyond classification requires the generalization of 
Theorem~\ref{thm:vapche}. For the  the square and absolute loss functions  and $\Y$ compact,
the characterization of learnability in terms of $\gamma$-fat shattering dimension can be used.
It is an open question whether there is a more direct way to show that learnability is sufficient for stability, independently to Theorem~\ref{thm:vapche} and to extend the above results to more general classes of loss functions. We will see a partial answer to this question in Section \ref{sec:GLS}.

\section{Learnability in the General Learning Setting}\label{sec:GLS}

In the previous sections we focused our attention on  supervised   learning. 
Here we ask  whether the results we discussed 
extend to the so called  general  learning \cite{Vap95}. 

Let $(\Z,\rho)$ be a probability space and $\F$  a measurable space. A loss function is a map
$L:\F\times\Z  \to [0,\infty)$, such that $L(f,\cdot)$ is measurable for all $f\in \F$. 
We are interested in the  problem of  minimizing the expected risk,
\beeq{eq:GLexp_risk}
{\inf_{\hh}\err,\quad\quad \err(f)=\int_{\Z} L(f,z)\,d\rho(z),}
when $\rho$ is fixed but known only through a training set,  ${\bf z}_n=(z_1,\ldots,z_n)\in \Z^n$ sampled  identically and independently  according  to $\rho$. 
Definition~\ref{Lalg} of a learning algorithm on $\hh$ applies as is to this setting and  ERM on $\hh$  is defined by the  minimization of  the empirical risk
$$
\empp(f)=\frac 1n \sum_{i=1}^n L(f,z_i).
$$
While   general learning  is close  to  supervised learning, there are
important differences.  The data space $\Z$ has no natural decomposition,  $\F$ needs not to be a space of functions.
Indeed,  $\F$ and  $\Z$ are related only via the loss function $L$.
For our discussion it is important to note that the distinction between $\F$ 
and the hypotheses space $\hh$  becomes blurred. 
In   supervised learning  $\F$ is  the largest  set of functions for which Problem~\eqref{eq:exp_risk}  is well defined (measurable functions in $\Y^\X$). The choice of a hypotheses corresponds intuitively to a  more "manageable" function space. 
In  general learning  the choice of $\F$ is more arbitrary  as a consequence the   the definition of 
universal hypotheses space is less clear. The setting is too general for  an analogue of the no free lunch theorem  to hold.
Given these premises,  in what follows we will simply identify $\F=\hh$ and consider the question of learnability, 
noting that the definition of uniform learnability extends naturally to  general learning. 
We  present  two sets of ideas.  The first, due to Vapnik,  focuses on a more restrictive notion of consistency of ERM. The second, investigates the characterization of uniform learnability in terms of stability.

\subsection{Vapnik's Approach and Non Trivial Consistency}

The extension of the classical results characterizing
learnability in terms of complexity measure is tricky. Since $\hh$ is not a function space
the definitions of $VC$ or $V_\gamma$ dimensions do not make sense.
A possibility is to  consider the class $L\circ \hh:=\{z\in\Z\mapsto L(f,z)\,\text{ for some } f\in\hh\}$ and the corresponding VC dimension (if $L$ is binary valued) or $V_\gamma$  dimension (if $L$ is real valued). Classic results about the  equivalence between the uGC property and finite complexity apply to the class  $L \circ \hh$. Moreover, uniform learnability can be easily proved if $L\circ \hh$ is a uGC class. On the contrary, the reverse implication does not hold in the general learning setting. A counterexample is given in \cite{Vap95} (Sec. 3.1)  showing that it is possible to design hypotheses classes with  infinite VC (or $V_\gamma$) dimension, which  are  uniformly learnable  with ERM.  The construction is as follows. Consider an arbitrary set $\hh$ and a loss $L$ for which the class $L\circ \hh$ has infinite VC (or $V_\gamma$) dimension. Define a new space $\widetilde{\hh}:=\hh\cup\{\tilde{h}\}$ by adding to $\hh$ an element $\tilde{h}$  such that $L(\tilde{h},z)\leq L(h,z)$ for all $z\in\Z$ and $h\in\hh$\footnote{Note that this construction is not possible in  classification or in  regression with the square loss.}\,. The space $L\circ \widetilde{\hh}$  has infinite VC, or $V_\gamma$, dimension and is trivially learnable by ERM, which is constant and coincides with $\tilde{h}$ for each probability measure $\rho$.  The previous counterexample proves that learnability, and in particular learnability via ERM, does not imply finite VC or $V_\gamma$ dimension.  
To avoid these cases of ``trivial consistency'' and to restore the equivalence between learnability and finite dimension, the following stronger notion of consistency for ERM has been introduced by Vapnik \cite{Vap95}.  
\begin{definition}  ERM on $\hh$   is {\em strictly uniformly consistent}  if and only if
\[
 \forall \epsilon>0,\quad
\lim_{n\to \infty} \sup_{\rho}\rho^n(
\,\inf_{\hh_c}\empp(f) -\inf_{\hh_c}\err(f)>\epsilon
)= 0,\]
where $\hh_c=\{f\in\hh\,:\, \err(f)\geq c\}$.
\end{definition}
The following result characterizes  strictly uniform consistency in terms of uGC property of the class $L\circ\hh$ (see Theorem 3.1 and its Corollary in \cite{Vap95}])
\begin{theorem}\label{thm:vap} 
Let $B>0$ and assume $L(f,z)\leq B$ for all $f\in\hh$ and $z\in \Z$. 
Then the following conditions are equivalent,
\begin{enumerate}
\item  ERM on $\hh$  is strictly consistent, 
\item $L\circ\hh$ is  a uniform one-sided  Glivenko-Cantelli class.
\end{enumerate}
\end{theorem}
The definition of one-sided Glivenko-Cantelli class simply corresponds to omitting  the absolute value in Definition \ref{def:uGC}. 

\subsection{Stability and  Learnability for General Learning}

In this section we discuss ideas from \cite{ShaShaSre10}  extending the stability 
approach to  general learning.   The following definitions are  relevant.
\begin{definition}\label{AERM}
A {\em uniform Asymptotic ERM (AERM) algorithm} $A$ on $\hh$ is a learning algorithm such that 
\[
\forall \epsilon >0, \quad
\lim_{n\to \infty} \,\sup_{\rho} \rho^n(\{\trs\,:\, \empp(A_{\trs})-\inf_{\hh}{\empp}>\epsilon \})=0.
\] 
\end{definition}
\begin{definition} A learning algorithm $A$ on $\hh$ is {\em uniformly  replace one (RO) stable} 
if there exists a sequence $\beta_n\to 0$ such that 
\[
\frac 1n \sum_{i=1}^n \vert L(A_{{\bf z}_n^{ i,u}},z)-L(A_{\trs},z)\vert \le \beta_n\,.
\]
for all $\trs\in \Z^n$, $u,z\in \Z^n$  and $i\in\{1,\ldots,n\}$.
\end{definition}
Note that the above definition is close to that of uniform stability~\eqref{ustab}, although the latter  turns out to be a stronger condition.  The importance of the above definitions is made clear by the following result. 
\begin{theorem}\label{thm:sha}
Let $B>0$ and assume $L(f,z)\leq B$ for all $f\in\hh$ and $z\in \Z$. 
Then the following conditions are equivalent,
\begin{enumerate}
\item $\hh$ is uniformly learnable,
\item there exists an  AERM algorithm  on $\hh$ which is  $RO$ stable.
\end{enumerate}
\end{theorem}
As mentioned in Remark \ref{rem:aerm}, Theorem \ref{stabtheo} holds not only for exact minimizers of the empirical risk, but also for AERM. In this view, there is a subtle difference between Theorem \ref{stabtheo} and Theorem \ref{thm:sha}. In supervised learning,  Theorem \ref{stabtheo} shows that  uniform learnability implies that {\em every}  ERM (AERM) is stable, while  in general learning, Theorem \ref{thm:sha} shows that uniform learnability implies the {\em existence} of a stable AERM (whose construction is not explicit).

The proof of the above result is given in Theorem 7 in \cite{ShaShaSre10}. The hard part of the proof is showing that learnability implies existence of a $RO$ stable AERM. 
This part of the proof is split in two steps  showing that: 1) if there is a uniformly consistent algorithm $A$, then there exists a uniformly consistent   AERM $A'$ (Lemma 20 and Theorem 10); 2) every uniformly consistent  AERM is also $RO$ stable (Theorem 9).
Note that the results in \cite{ShaShaSre10} are given in expectation and with some 
quantification of how different convergence rates are related. Here we give results in probability to be uniform with the rest of the paper and state only asymptotic results 
to simplify the presentation.

\section{Discussion}
In this paper we reviewed  several results concerning    learnability of a hypotheses space. 
Extensions of these ideas can be found in  \cite{DS11} (and references therein) for  multi-category classification, and in \cite{RST11} 
for sequential prediction.    It would be interesting to devise constructive proofs in general learning  suggesting how stable learning algorithms can be designed. Moreover, it would be interesting to study universal consistency and learnability in the case of samples from non stationary processes.
%

%
%
%
%
%
\bibliographystyle{plain}
\bibliography{tutta_la_bib}

\begin{thebibliography}{10}

\bibitem{AloBenCes97}
N.~Alon, S.~Ben-David, N.~Cesa-Bianchi, and D.~Haussler.
\newblock Scale-sensitive dimensions, uniform convergence, and learnability.
\newblock {\em J. ACM}, 44(4):615--631, 1997.

\bibitem{AntBar99}
M.~Anthony and P.~L. Bartlett.
\newblock {\em Neural network learning: theoretical foundations}.
\newblock Cambridge University Press, Cambridge, 1999.

\bibitem{BarLonWil96}
P.~Bartlett, P.~Long, and R.~Williamson.
\newblock Fat-shattering and the learnability of real-valued functions.
\newblock {\em Journal of Computer and System Sciences}, 52:434--452, 1996.

\bibitem{bousquet02stability}
O.~Bousquet and A.~Elisseeff.
\newblock Stability and generalization,.
\newblock {\em Journal of Machine Learning Research}, 2:499--526, 2002.

\bibitem{DS11}
Amit Daniely, Sivan Sabato, Shai Ben-David, and Shai Shalev-Shwartz.
\newblock Multiclass learnability and the erm principle.
\newblock {\em Journal of Machine Learning Research - Proceedings Track},
  19:207--232, 2011.

\bibitem{DevGyoLug96}
L.~Devroye, L.~Gy\"orfi, and G.~Lugosi.
\newblock {\em A Probabilistic Theory of Pattern Recognition}.
\newblock Number~31 in Applications of mathematics. Springer, New York, 1996.

\bibitem{DudGinZin91}
R.~Dudley, E.~Gin\'e, and J.~Zinn.
\newblock Uniform and universal {G}livenko-{C}antelli classes.
\newblock {\em J. Theoret. Prob.}, 4:485--510, 1991.

\bibitem{enhane}
H.~W. Engl, M.~Hanke, and A.~Neubauer.
\newblock {\em Regularization of inverse problems}, volume 375 of {\em
  Mathematics and its Applications}.
\newblock Kluwer Academic Publishers Group, Dordrecht, 1996.

\bibitem{KeaSch94}
Michael~J. Kearns and Robert~E. Schapire.
\newblock Efficient distribution-free learning of probabilistic concepts.
\newblock In {\em Computational learning theory and natural learning systems,
  {V}ol.\ {I}}, Bradford Book, pages 289--329. MIT Press, Cambridge, MA, 1994.

\bibitem{KutNiy02}
S.~Kutin and P.~Niyogi.
\newblock Almost-everywhere algorithmic stability and generalization error.
\newblock Technical Report TR-2002-03, Department of Computer Science, The
  University of Chicago, 2002.

\bibitem{MukNyiPog06}
S.~Mukherjee, P.~Niyogi, T.~Poggio, and R.~Rifkin.
\newblock Learning theory: stability is sufficient for generalization and
  necessary and sufficient for consistency of empirical risk minimization.
\newblock {\em Adv. Comput. Math.}, 25(1-3):161--193, 2006.

\bibitem{PogRifMuk04}
T.~Poggio, R.~Rifkin, S.~Mukherjee, and P.~Niyogi.
\newblock General conditions for predictivity in learning theory.
\newblock {\em Nature}, 428:419--422, 2004.

\bibitem{RST11}
Alexander Rakhlin, Karthik Sridharan, and Ambuj Tewari.
\newblock Online learning: Beyond regret.
\newblock {\em Journal of Machine Learning Research - Proceedings Track},
  19:559--594, 2011.

\bibitem{ShaShaSre10}
S.~Shalev-Shwartz, O.~Shamir, N.~Srebro, and K.~Sridharan.
\newblock Learnability, stability and uniform convergence.
\newblock {\em J. Mach. Learn. Res.}, 11:2635--2670, 2010.

\bibitem{SteChr08}
I.~Steinwart and A.~Christmann.
\newblock {\em Support vector machines}.
\newblock Information Science and Statistics. Springer, New York, 2008.

\bibitem{Vap95}
V.~Vapnik.
\newblock {\em The Nature of Statistical Learning Theory}.
\newblock Springer Verlag, New York, 1995.

\bibitem{VapChe71}
V.~N. Vapnik and A.~Y. Chervonenkis.
\newblock Theory of uniform convergence of frequencies of events to their
  probabilities and problems of search for an optimal solution from empirical
  data.
\newblock {\em Avtomat. i Telemeh.}, (2):42--53, 1971.

\end{thebibliography}
\end{document}